\documentclass[runningheads]{llncs}
\usepackage[T1]{fontenc}
\usepackage{graphicx,verbatim}
\usepackage{xcolor}
\usepackage{amssymb}
\usepackage{amsmath}
\usepackage{booktabs} 

\usepackage[table]{xcolor}
\usepackage[usenames,dvipsnames]{xcolor}
\usepackage{pgfplots}
\usepgfplotslibrary{groupplots}
\pgfplotsset{compat=1.18}

\usepackage{subcaption}
\usepackage{tabularx}
\usepackage{hyperref}
\usepackage{placeins}  % in the preamble
\definecolor{myred}{HTML}{FF595E}
\definecolor{mygreen}{HTML}{8AC926}
\definecolor{myblue}{HTML}{1982C3}
\definecolor{mygray}{HTML}{797D81}
\definecolor{mypurple}{HTML}{694C93}
\definecolor{myyellow}{HTML}{FFCA3A}
\definecolor{mypink}{HTML}{FF99D3}
\definecolor{myaqua}{HTML}{35D4C4}
\definecolor{tablerow}{RGB}{238, 238, 238}
\begin{document}
\title{Whole-Body MRI Classification via Prompt-Based Clinical Conditioning}

\author{Laura Daza\inst{1,2}\orcidID{0000-0003-4170-6168} \and
Marta Hasny\inst{1,2}\orcidID{0000-0001-5634-1240} \and
Cristina González\inst{1,2}\orcidID{0000-0001-9445-9952} \and
Julia A. Schnabel\inst{1,2,3,4}\orcidID{0000-0001-6107-3009}}
%

%index{Daza, Laura}
%index{Hasny, Marta}
%index{González, Cristina}
%index{Schnabel, Julia A}

\authorrunning{L. Daza et al.}
\institute{Institute of Machine Learning in Biomedical Imaging, Helmholtz Munich, Germany \and
School of Computation, Information and Technology, Technical University of Munich, Germany \and
School of Biomedical Engineering \& Imaging Sciences, King’s College London, UK \and Munich Center for Machine Learning, Germany}

\maketitle              % typeset the header of the contribution
\begin{abstract}

Combining whole-body magnetic resonance imaging (WB-MRI) with clinical variables has the potential to improve systemic disease diagnosis by leveraging complementary sources of patient information. However, structured clinical variables are often incomplete or missing, limiting the applicability of conventional multimodal fusion methods that assume fixed inputs. In this work, we propose TACTIC (Tabular-Attribute Conditioned Transformer for Image Classification), a prompt-based multimodal framework that integrates WB-MRI and structured clinical data through conditional visual feature learning. By encoding clinical attributes as prompts, TACTIC supports an arbitrary number of tabular inputs and naturally handles missing data without requiring imputation or fixed input structures.
We evaluate TACTIC on five WB-MRI classification tasks spanning systemic and oncologic applications, including diabetes, chronic obstructive pulmonary disease (COPD), breast cancer, prostate cancer, and metastasis diagnosis. Across all tasks, TACTIC consistently improves performance over image-only baselines when clinical information is available while maintaining strong predictive capability under incomplete tabular inputs.
Our results demonstrate the effectiveness of prompt-based models as a flexible approach for improving WB-MRI analysis using clinical context. The model weights and code are available at \url{https://github.com/lauradaza/TACTIC}. 

\keywords{Whole body MRI \and Tabular clinical data \and Clinical prompting \and Disease classification}
% Authors must provide keywords and are not allowed to remove this Keyword section.

\end{abstract}
\section{Introduction}

Whole-body magnetic resonance imaging (WB-MRI) offers a comprehensive view of the human body by simultaneously capturing multiple organs and anatomical systems without ionizing radiation, making it particularly attractive for systemic disease assessment and population-scale screening. Recent studies have demonstrated the potential of combining WB-MRI with clinical data for preclinical disease risk assessment~\cite{seletkov2025ai}. Despite these advantages and its established role in the management of several diseases~\cite{summers2021whole,pflugfelder2013malignant,messiou2019guidelines,mottet2017eau}, WB-MRI remains underutilized in automated diagnosis pipelines.

In clinical practice, diagnostic and prognostic decisions are inherently multimodal, relying on imaging findings and structured clinical variables that are not directly observable from imaging data, such as laboratory test results, genetic predispositions, and medical history. Integrating these complementary sources of information is therefore essential to fully leverage imaging data for personalized medicine. Recent large-scale datasets, such as the UK Biobank~\cite{sudlow2015uk}, provide a unique opportunity to study this integration by combining WB-MRI with extensive clinical and demographic data. Early studies have demonstrated the promise of such approaches for medical applications~\cite{seletkov2025ai,du2024tip,hasny2025no}, yet the development of robust multimodal learning strategies remains an active challenge.

A key limitation in current multimodal methods is their assumption of complete and consistently available clinical data. In practice, clinical variables are often missing, sparsely recorded, or heterogeneous across patients, even in well-curated cohorts. This systematic missingness poses a significant obstacle for conventional fusion techniques, which typically rely on fixed-dimensional tabular inputs~\cite{vale2021long,spasov2019parameter}. Consequently, their performance degrades when the availability of clinical information varies, limiting their applicability in real-world settings.

Recent advances in prompt-based vision models allow flexible conditioning of model behavior through external inputs~\cite{kirillov2023segment,ma2024segment}, enabling the integration of auxiliary information without requiring a fixed number of prompts. In the medical domain, such approaches provide a promising mechanism for incorporating patient-specific context in an adaptive manner. However, existing approaches predominantly rely on spatial cues~\cite{isensee2025nninteractive,wang2025sam}, free-form textual descriptions from reports~\cite{zhang2023biomedclip} or translated from structured clinical variables~\cite{milecki2023medimp,huang2025bridging}, as well as prompt-learning strategies over multimodal inputs~\cite{kang2023visual}. As a result, they do not naturally leverage structured clinical data in its native form, often requiring additional preprocessing or modality-specific adaptation.

In this work, we propose TACTIC (Tabular-Attribute Conditioned Transformer for Image Classification), a prompt-based framework designed to leverage WB-MRI with complementary clinical data for disease classification. TACTIC reformulates tabular clinical variables in their native form as prompts that complement image representations, allowing patient-specific information to directly guide visual feature learning. This design enables the model to flexibly incorporate an arbitrary number of clinical variables, making it inherently robust to missing or partially observed data.
We evaluate TACTIC on multiple WB-MRI classification tasks spanning systemic and oncologic applications, including diabetes, chronic obstructive pulmonary disease (COPD), breast cancer, prostate cancer, and metastasis diagnosis. Across all tasks, TACTIC consistently improves performance over image-only models when clinical data is available, while maintaining strong predictive capability under limited or missing tabular inputs.

Our contributions are threefold: (i) we introduce TACTIC, a prompt-based multimodal framework that conditions WB-MRI representations on structured clinical attributes; (ii) we reformulate multimodal fusion as a conditioning problem, enabling flexible integration of heterogeneous and incomplete clinical data with pretrained visual backbones; and (iii) we demonstrate the applicability of TACTIC across diverse WB-MRI classification tasks, highlighting the potential of WB-MRI as a foundation for clinically relevant multimodal disease prediction.

\section{Method}

\subsection{Tabular-Attribute Conditioned Transformer}
\begin{figure}[t]
    \centering
    \includegraphics[width=\linewidth]{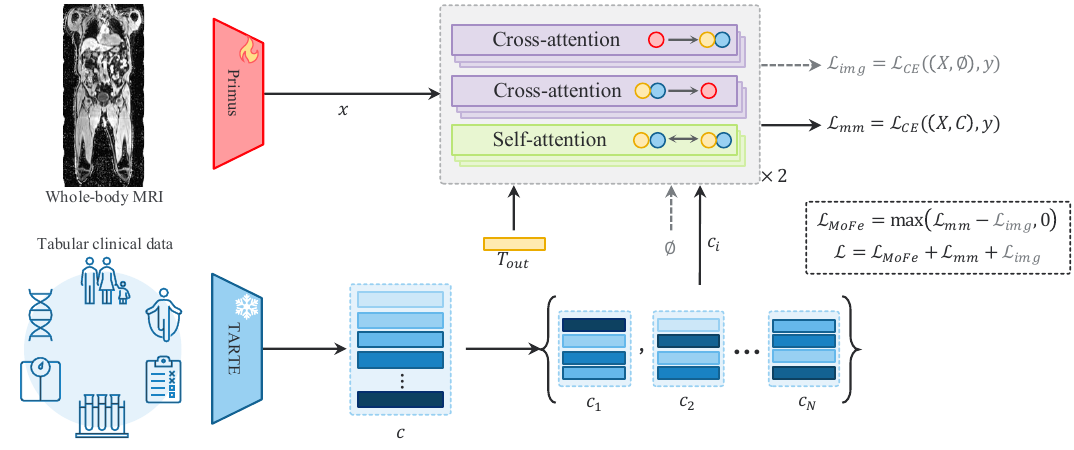}
    \caption{Overview of TACTIC. The WB-MRI features $x$ are conditioned using random subsets $c_i$ of clinical features as prompts. Combined with an output token $T_{\text{out}}$, the prompts guide image feature aggregation through cross-attention, enabling flexible multimodal fusion under variable clinical data availability. WB-MRI reproduced by kind permission of UK Biobank ©}
    \label{fig:overview}
\end{figure}

As shown in Figure~\ref{fig:overview}, our method models structured clinical variables as prompts that condition the image classification process, enabling patient-specific adaptation of visual representations. We adopt Primus-M~\cite{wald2025primus} as the image encoder and use TARTE~\cite{kim2025table} to embed structured clinical information. Specifically, TARTE generates an embedding for each clinical attribute by jointly encoding the attribute name (e.g., \textit{sex}, \textit{age}) and its corresponding value (e.g., \textit{female}, \textit{64}).

The core of our approach is a prompt-conditioned transformer that integrates clinical and visual information. TACTIC first performs self-attention over the clinical prompts to capture global dependencies among tabular attributes. These refined prompts then interact with image features through cross-attention, guiding the aggregation of visual information in a patient-specific manner. By treating clinical context as prompts rather than as additional input modalities, TACTIC enables a simple integration with pretrained visual backbones, while injecting clinically relevant information for accurate disease classification.

\subsection{Single Modality downstream training}

\subsubsection{Image Encoder.}

We pretrain Primus using a masked autoencoding (MAE)~\cite{he2022masked} framework adapted for WB-MRI. Since more than 40\% of the voxels in WB-MRI scans are outside the body, we first generate body masks using otsu thresholding~\cite{otsu1979threshold} and binary morphology to identify foreground regions. These masks are used in two ways. First, we apply a mask-aware pooling operation that merges voxels in windows of size $[2\times2\times4]$ while ignoring positions outside the body, thereby reducing the image size with minimal information loss. Second, even after pooling, a substantial fraction of the resulting tokens still correspond primarily to background voxels, biasing the standard MAE reconstruction loss toward trivial background patches. Therefore, we restrict masking and reconstruction to patches containing anatomical information, ensuring that the pretraining objective focuses on relevant content and promotes the learning of meaningful image representations.

\subsubsection{Tabular Encoder.}
We train TARTE by explicitly simulating missing clinical data, following~\cite{hasny2025no}. Let $C = {c^{(1)}, \ldots, c^{(K)}}$ denote the full set of tabular attributes. During training, we sample a subset $C_{\text{more}} \subseteq C$ and a second subset $C_{\text{few}} \subset C_{\text{more}}$, representing scenarios with more and fewer available clinical attributes, respectively. Predictions obtained from both subsets are used to compute the Tabular More vs.\ Fewer (TabMoFe) loss~\cite{hasny2025no}, which encourages improved performance as additional clinical information becomes available:

\begin{equation}
\mathcal{L}_{\text{TabMoFe}} = \max\bigl(\mathcal{L}_{\text{more}} - \mathcal{L}_{\text{few}}, 0\bigr)
\end{equation}

The final tabular loss is given by \(\mathcal{L}_{\text{tab}} = \mathcal{L}_{\text{TabMoFe}} + \mathcal{L}_{\text{more}} + \mathcal{L}_{\text{few}}\), where $\mathcal{L}_{\text{more}}$ and $\mathcal{L}_{\text{few}}$ denote the cross entropy losses obtained using $C_{\text{more}}$ and $C_{\text{few}}$.

\subsection{Clinical Prompting Training}

Given paired imaging and tabular clinical data $(X, C)$ with diagnosis label $y$, we first extract unimodal representations using their corresponding encoders. The image encoder produces image features $x \in \mathbb{R}^{n \times d}$, while the tabular encoder produces clinical embeddings $c \in \mathbb{R}^{K \times d}$.
To simulate variable clinical data availability, we sample $N$ random subsets of clinical attributes $\{c_1, c_2, \ldots, c_N\}$, where each subset $c_i \subseteq C$ contains $k$ attributes. Each subset is concatenated with an output token $T_{\text{out}}$ and treated as a set of prompts that, together with the image features $x$, condition the model's output. A multimodal prediction is obtained for each subset, and the multimodal loss $\mathcal{L}_{\text{mm}}$ is computed using the average prediction across all sampled subsets.

In parallel, we compute an image-only loss $\mathcal{L}_{\text{img}}$ by conditioning the model with the output token $T_{\text{out}}$ and an empty clinical prompt set $\varnothing$. The model is trained using the More vs.\ Fewer (MoFe) loss~\cite{li2025simmlm}, defined as
\begin{equation}
\mathcal{L}_{\text{MoFe}} = \max\bigl(\mathcal{L}_{\text{mm}} - \mathcal{L}_{\text{img}}, 0\bigr)
\end{equation}
where $\mathcal{L}_{\text{mm}} = \mathcal{L}_{CE}((X, C), y)$, $\mathcal{L}_{\text{img}} = \mathcal{L}_{CE}((X, \varnothing), y)$, and $\mathcal{L}_{CE}$ denotes the cross-entropy loss. The final training objective is \(\mathcal{L} = \mathcal{L}_{\text{MoFe}} + \mathcal{L}_{\text{mm}} + \mathcal{L}_{\text{img}}\).
\section{Experimental Setting}

\subsection{Dataset}

The UK Biobank~\cite{sudlow2015uk} contains over 500,000 participants, with a subset of 80,000 that includes WB-MRI data and corresponding clinical information from in hospital inpatient records. WB-MRI are neck-to-knee T1-weighted dual-echo Dixon acquisitions with a spatial resolution of $[2.23 \times 2.23 \times 3.0]$ mm and size $[224 \times 168 \times 363]$. For all experiments, we use the fat-only and water-only image reconstructions. In addition to imaging, we extract 235 clinical attributes following~\cite{lubeck2025adaptable}, including demographic information, medical and family history, blood and urine biomarkers, polygenic risk scores, lifestyle factors, physical measurements, and sex-specific variables.

\begin{table}[t]
\centering
    \caption{Training subset sizes per disease. The sex distribution of the patients is maintained between the healthy and diagnosed subjects.}
    \begin{tabularx}{\textwidth}{
    l
    >{\centering\arraybackslash}X
    >{\centering\arraybackslash}X
    >{\centering\arraybackslash}X
    >{\centering\arraybackslash}X
    >{\centering\arraybackslash}X
    }

    \toprule
    & Diabetes & COPD & Breast cancer & Prostate cancer & Metastasis \\
    \midrule
    Subset size & 4060 & 1292 & 3092 & 2682 & 2885 \\
    \% Female & 35.5\% & 45.2\% & 99.4\% & 0.0\% & 88.7\% \\
    \bottomrule

\end{tabularx}
\label{tab:dataset}
\end{table}

For disease classification we use labels derived from the International Classification of Diseases (ICD-10) codes. We evaluate our method on five classification tasks: (1) diabetes, (2) chronic obstructive pulmonary disease (COPD), (3) breast cancer, (4) prostate cancer, and (5) metastasis in our cancer cohorts. We include participants with imaging and clinical information and split them into training, validation, and test sets using a 70–10–20 ratio. As all target conditions are under-represented, we construct balanced training subsets for all downstream tasks. For these subsets, participants without the target condition are randomly sampled ensuring the preservation of the sex distribution. For metastasis classification, we use our full cancer cohort, which has a prevalence of 11\% metastatic cases. The resulting subset sizes are reported in Table~\ref{tab:dataset}.

\subsection{Implementation Details}

We pretrain Primus using all training WB-MRI scans of the UK Biobank. The pretraining is done for $50$ epochs using input images of size $[224 \times 160 \times 352]$, masking ratio of 90\%, learning rate of $8\times10^{-4}$, and two warm-up epochs followed by cosine annealing.
During fine-tuning, we use the same architecture and training configuration for all tasks. For tabular fine-tuning, we use a learning rate of $1\times10^{-5}$, batch size of $128$, and train for $100$ epochs. Imaging and~multimodal tuning are done with an image masking rate of 40\% to eliminate background tokens and reduce computational costs, batch size of $48$ and gradient accumulation over $6$ steps, resulting in an effective batch of $288$. We initialize TACTIC from the fine-tuned unimodal model weights and optimize until convergence with TARTE frozen. All experiments are performed in one NVIDIA H100 GPU with 80Gb of RAM.

\section{Results}
\vspace{-4pt}
We compare TACTIC against single- and multimodal baselines. The single-modality models are the vision-only Primus and the tabular-only TARTE. For multimodal fusion, we evaluate Max Fuse~\cite{vale2021long}, Concat Fuse~\cite{spasov2019parameter}, Sum Fuse, DAFT~\cite{Poelsterl2021-daft}, Dynamic Mixture of Modality Experts (DMoMe)~\cite{li2025simmlm}, and TIP~\cite{du2024tip} under varying levels of clinical data missingness. Performance is measured using the area under the receiver operating characteristic curve (AUROC), reported as the mean across ten levels of tabular data availability from 10\% to 100\%.

All multimodal methods are initialized from the same pretrained image and tabular encoders as TACTIC and trained for an identical number of epochs to ensure a fair comparison. For Max Fuse and Concat Fuse, we follow the original formulations and apply a linear classifier to the fused embeddings. We further include a Sum Fuse baseline that combines modality embeddings via element-wise addition. For DAFT, a DAFT block is inserted before the final transformer block of the image backbone. For DMoMe, we adopt the mixture-of-experts variant proposed for UPMC Food-101~\cite{allegra2017multimedia}, using intermediate feature representations for gating to mitigate the computational cost of processing raw WB-MRI data. Finally, the substantially larger size of WB-MRI volumes compared with the inputs originally used in TIP makes its pretraining stage computationally impractical. We therefore adopt the original downstream fine-tuning framework developed for cardiac imaging and replace only the unimodal backbone architectures with our image and tabular encoders.

\subsection{Performance Comparison}
\begin{table}[t]
\centering
    \caption{Performance of TACTIC compared to multiple methods on 5 classification tasks in the UK Biobank. We report the mean AUROC and the improvement over Primus under varying levels of tabular data availability (10\% - 100\%)}
    \newcommand{\Good}[2]{%
  #1 {\scriptsize \textcolor{ForestGreen}{+#2}}%
}
\newcommand{\Bad}[2]{%
  #1 {\scriptsize \textcolor{red}{ - #2}}%
}

\begin{tabularx}{\textwidth}{
    l
    >{\centering\arraybackslash}X
    >{\centering\arraybackslash}X
    >{\centering\arraybackslash}X
    >{\centering\arraybackslash}X
    >{\centering\arraybackslash}X
    }

    \toprule
    Method & Diabetes & COPD & Breast cancer & Prostate cancer & Metastasis \\
    \midrule
    \rowcolor{tablerow} \textit{Single-modality} & & & & & \\
    Primus (img)~\cite{wald2025primus} & 78.8\phantom{\scriptsize + 0.00} & 74.0\phantom{\scriptsize + 0.0} & 58.0\phantom{\scriptsize + 0.0} & 49.0\phantom{\scriptsize + 0.00} & 66.1\phantom{\scriptsize + 0.0} \\
    TARTE (tab)~\cite{kim2025table} & 87.1\phantom{\scriptsize + 0.00} & 79.3\phantom{\scriptsize + 0.0} & 62.8\phantom{\scriptsize + 0.0} & \textbf{78.2\phantom{\scriptsize + 0.00}} & 64.7\phantom{\scriptsize + 0.0} \\
    \midrule
    \rowcolor{tablerow} \textit{Multimodal fusion} & & & & & \\
    Concat Fuse~\cite{spasov2019parameter} & \Good{89.1}{10.3} & \Good{80.1}{6.1} & \Good{61.2}{3.2} & \Good{77.8}{28.8} & \Bad{66.0}{0.2} \\
    Max Fuse~\cite{vale2021long} & \Good{89.0}{10.2} & \Good{77.8}{3.8} & \Good{58.7}{0.7} & \Good{75.2}{26.2} & \Good{67.1}{1.0} \\
    Sum Fuse & \Good{88.7}{ 9.9} & \Good{79.2}{5.2} & \Good{59.5}{1.5} & \Good{77.4}{28.5} & \Good{66.7}{0.5} \\
    DAFT~\cite{Poelsterl2021-daft} & \Good{78.9}{ 0.1} & \Bad{66.6}{7.4} & \Bad{45.9}{12.1} & \Good{52.3}{ 3.4} & \Bad{64.8}{1.3} \\
    DMoMe~\cite{li2025simmlm} & \Good{89.0}{10.2} & \Good{76.3}{2.3} & \Good{61.0}{3.0} & \Good{74.1}{25.2} & \Good{67.6}{1.5} \\
    TIP~\cite{du2024tip} & \Good{82.8}{ 4.0} & \Good{76.5}{2.5} & \Good{61.0}{3.0} & \Good{73.2}{24.2} & \Bad{62.7}{3.5} \\
    % \midrule
    TACTIC (ours) & \textbf{\Good{89.1}{10.3}} & \textbf{\Good{83.3}{9.3}} & \textbf{\Good{63.9}{5.9}} & \Good{76.6}{27.7} & \textbf{\Good{69.2}{3.1}} \\
    \bottomrule

\end{tabularx}
\label{tab:main_results}
\end{table}

Table~\ref{tab:main_results} shows that TACTIC consistently improves upon the image-only baseline across all classification tasks and achieves the best overall performance for four out of five diseases. For diabetes, clinical variables are highly predictive on their own, achieving an AUROC of 87.1. Consequently, all multimodal methods substantially outperform the image-only baseline. TACTIC matches the best-performing fusion approach, demonstrating that conditioning image representations with clinical prompts can effectively exploit the complementary information present in both modalities.
For COPD and breast cancer, the performance gap between the image- and tabular-only models is smaller, indicating that neither modality alone is sufficient for optimal prediction. In these settings, conventional fusion approaches often fail to effectively combine both sources of information and may even degrade performance relative to the tabular-only baseline. In contrast, TACTIC achieves the highest AUROC for both tasks, surpassing both single modality models and highlighting its ability to capture complementary information from heterogeneous modalities.

Prostate cancer shows a different pattern, with substantially lower performance for the image-only model than for the tabular baseline, suggesting that WB-MRI provides limited discriminative information for this task. As a result, most multimodal fusion methods highly benefit from incorporating clinical data, with Concat Fuse archiving the best performance. While TACTIC also benefits from clinical conditioning, we start from image representations and inject clinical context through prompting, which leads to smaller performance gains in scenarios where imaging contributes minimal task-relevant information.
For metastasis detection, the image-only model outperforms the tabular-only baseline, indicating that WB-MRI is the main source of predictive information. In this scenario, some fusion methods fail to improve upon the image-only baseline, suggesting that in this task the clinical variables can introduce noise. In contrast, TACTIC achieves the largest improvement over the image-only model, demonstrating that prompt-based conditioning can effectively leverage complementary patient information without compromising the visual representations.

\subsection{Incomplete Clinical Data}

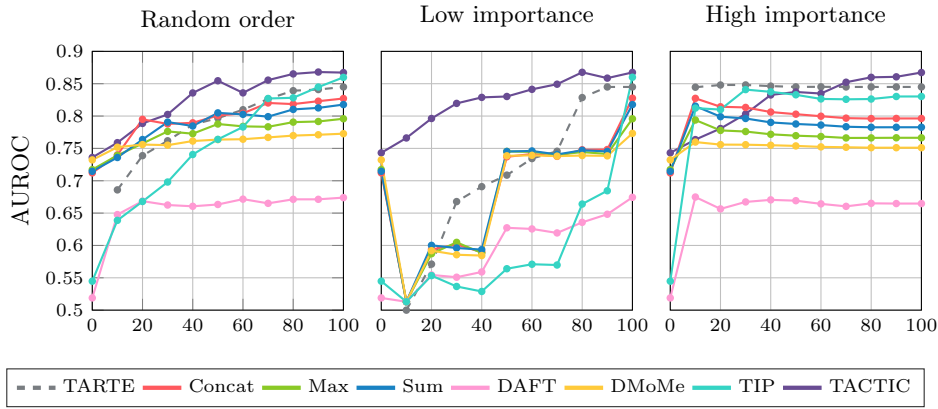
\begin{figure}[t!]
    \centering
    \begin{tikzpicture}

\begin{groupplot}[
    group style={
        group size=3 by 1,
        horizontal sep=0.5cm,
        ylabels at=edge left,
    },
    width=4.9cm,
    height=5cm,
    xmin=0, xmax=100,
    ymin=0.5, ymax=0.9,
    ytick={0.5,0.55,0.6,...,0.95},
    grid=both,
    tick label style={font=\scriptsize},
    title style={font=\small},
    y label style={font=\small},
]

% --- plots here ---
% --- RANDOM ---
\nextgroupplot[title={Random order}, ylabel={AUROC}]
\addplot[color=mygray, dashed, thick, mark=*, mark size=1pt, mark options={solid}] table[x=Available,y=Tabular]{figures/plots_data/copd.txt};
\addplot[color=mypurple, thick, mark=*, mark size=1pt, mark options={solid}] table[x=Available,y=TACTIC]{figures/plots_data/copd.txt};
\addplot[color=myred, thick, mark=*, mark size=1pt, mark options={solid}] table[x=Available,y=Concat]{figures/plots_data/copd.txt};
\addplot[color=mygreen, thick, mark=*, mark size=1pt, mark options={solid}] table[x=Available,y=Max]{figures/plots_data/copd.txt};
\addplot[color=myblue, thick, mark=*, mark size=1pt, mark options={solid}] table[x=Available,y=Sum]{figures/plots_data/copd.txt};
\addplot[color=myyellow, thick, mark=*, mark size=1pt, mark options={solid}] table[x=Available,y=Gate]{figures/plots_data/copd.txt};
\addplot[color=mypink, thick, mark=*, mark size=1pt, mark options={solid}] table[x=Available,y=DAFT]{figures/plots_data/copd.txt};
\addplot[color=myaqua, thick, mark=*, mark size=1pt, mark options={solid}] table[x=Available,y=TIP]{figures/plots_data/copd.txt};

\nextgroupplot[title={Low importance}, yticklabel=\empty, tick style={draw=none}]
\addplot[color=mygray, dashed, thick, mark=*, mark size=1pt, mark options={solid}] table[x=Available,y=Tabular]{figures/plots_data/LI.txt};
\addplot[color=mypurple, thick, mark=*, mark size=1pt, mark options={solid}] table[x=Available,y=TACTIC]{figures/plots_data/LI.txt};
\addplot[color=myred, thick, mark=*, mark size=1pt, mark options={solid}] table[x=Available,y=Concat]{figures/plots_data/LI.txt};
\addplot[color=mygreen, thick, mark=*, mark size=1pt, mark options={solid}] table[x=Available,y=Max]{figures/plots_data/LI.txt};
\addplot[color=myblue, thick, mark=*, mark size=1pt, mark options={solid}] table[x=Available,y=Sum]{figures/plots_data/LI.txt};
\addplot[color=myyellow, thick, mark=*, mark size=1pt, mark options={solid}] table[x=Available,y=Gate]{figures/plots_data/LI.txt};
\addplot[color=mypink, thick, mark=*, mark size=1pt, mark options={solid}] table[x=Available,y=DAFT]{figures/plots_data/LI.txt};
\addplot[color=myaqua, thick, mark=*, mark size=1pt, mark options={solid}] table[x=Available,y=TIP]{figures/plots_data/LI.txt};

\nextgroupplot[title={High importance}, yticklabel=\empty, tick style={draw=none}]
\addplot[color=mygray, dashed, thick, mark=*, mark size=1pt, mark options={solid}] table[x=Available,y=Tabular]{figures/plots_data/HI.txt};
\addplot[color=mypurple, thick, mark=*, mark size=1pt, mark options={solid}] table[x=Available,y=TACTIC]{figures/plots_data/HI.txt};
\addplot[color=myred, thick, mark=*, mark size=1pt, mark options={solid}] table[x=Available,y=Concat]{figures/plots_data/HI.txt};
\addplot[color=mygreen, thick, mark=*, mark size=1pt, mark options={solid}] table[x=Available,y=Max]{figures/plots_data/HI.txt};
\addplot[color=myblue, thick, mark=*, mark size=1pt, mark options={solid}] table[x=Available,y=Sum]{figures/plots_data/HI.txt};
\addplot[color=myyellow, thick, mark=*, mark size=1pt, mark options={solid}] table[x=Available,y=Gate]{figures/plots_data/HI.txt};
\addplot[color=mypink, thick, mark=*, mark size=1pt, mark options={solid}] table[x=Available,y=DAFT]{figures/plots_data/HI.txt};
\addplot[color=myaqua, thick, mark=*, mark size=1pt, mark options={solid}] table[x=Available,y=TIP]{figures/plots_data/HI.txt};

\end{groupplot}

\end{tikzpicture}

\begin{center}
\begin{tikzpicture}
\begin{axis}[
    hide axis,
    xmin=0, xmax=1, ymin=0, ymax=1,
    width=\textwidth,
    height=2cm,
    legend style={
        draw=white!15!black, % Added a slight border for visibility
        font=\scriptsize,
        legend columns=8,
    },
    legend cell align={center},
    legend image post style={xscale=0.85}] % shorter legend lines

    \addlegendimage{mygray, ultra thick, dashed}
    \addlegendentry{TARTE}  % ~\cite{kim2025table}

    \addlegendimage{myred, ultra thick}
    \addlegendentry{Concat}  % ~\cite{spasov2019parameter}
    
    \addlegendimage{mygreen, ultra thick}
    \addlegendentry{Max}  % ~\cite{vale2021long}
    
    \addlegendimage{myblue, ultra thick}
    \addlegendentry{Sum}
    
    \addlegendimage{mypink, ultra thick}
    \addlegendentry{DAFT}  % ~\cite{Poelsterl2021-daft}

    \addlegendimage{myyellow, ultra thick}
    \addlegendentry{DMoMe}  % ~\cite{li2025simmlm}
    
    \addlegendimage{myaqua, ultra thick}
    \addlegendentry{TIP}  % ~\cite{}
    
    \addlegendimage{mypurple, ultra thick}
    \addlegendentry{TACTIC}
    
\end{axis}
\end{tikzpicture}
\end{center}
    \caption{Classification performance under incomplete clinical data for COPD. Results are reported as a function of the fraction of available clinical attributes, sampled randomly, in ascending and in descending order of attribute importance}
    \label{fig:plots}
\end{figure}

Figure~\ref{fig:plots} reports model performance under varying levels of clinical data availability for COPD. We progressively increase the fraction of visible clinical attributes using three sampling strategies: random selection, ascending, and descending order of attribute importance. Attribute importance is estimated from TARTE's attention weights over the clinical variables on the training set.
Across all scenarios, TACTIC consistently achieves the highest performance and shows the strongest robustness to incomplete clinical information. 

Under random attribute selection, TACTIC outperforms all multimodal approaches over the entire range of attribute availability, even when only a small subset of clinical variables is provided. As more attributes become available, its performance steadily improves, indicating effective integration of complementary information from both modalities.
When only low-importance attributes are available, most fusion methods experience a performance drop until a large fraction of variables becomes available. In contrast, TACTIC is always better than the image-only starting point. 
When we start from the high-importance attributes, all multimodal methods benefit from the initial tabular data. However, as less informative information is introduced, the performance of most methods degrades, while TACTIC consistently improves. Together, these results suggest that our prompt-based approach can leverage all the available clinical information without affecting the visual representations with noisy or less informative clinical data.
Overall, our results demonstrate that prompt-based conditioning provides a reliable mechanism for integrating clinical information.

\section{Conclusion}

In this work we introduce TACTIC, a prompt-based multimodal framework that complements WB-MRI with structured clinical variables for disease classification. By conditioning image representations on clinical attributes prompts, our method provides a flexible mechanism to incorporate patient-specific context without relying on fixed or complete tabular inputs. This design directly addresses a common limitation of conventional multimodal fusion strategies, which often degrade performance when clinical data is incomplete or fully missing.
Across a range of systemic and oncological tasks, we showed that our model consistently improves upon image-only models when clinical information is available, and remains effective even when only a limited subset of clinical variables is available. Compared to previous fusion approaches, TACTIC demonstrates greater robustness under incomplete clinical data settings and more stable performance as the amount of available tabular information varies. These results suggest that conditioning visual representations via prompting enables more effective exploitation of complementary clinical cues than direct feature aggregation.
Overall, our findings highlight the potential of flexible multimodal frameworks to enhance WB-MRI analysis in clinically realistic settings, particularly in scenarios where clinical data availability is inherently heterogeneous.

\begin{credits}
\subsubsection{\ackname} This research has been conducted using the UK Biobank Resource under Application Number 87065. LD and MH are supported by the German Federal Ministry of Research, Technology and Space (DECIPHER-M, 01KD2420G). MH is in part supported by the Munich School of Data Science (MUDS) and the European Laboratory for Learning and Intelligent Systems (ELLIS) PhD program. 

\subsubsection{\discintname}
The authors have no competing interests to declare that are
relevant to the content of this article.
\end{credits}

\bibliographystyle{splncs04}
\bibliography{mybibliography}

\end{document}